\documentclass[conference]{IEEEtran}
\IEEEoverridecommandlockouts
\usepackage{cite}
\usepackage{amsmath,amssymb,amsfonts}
\usepackage{algorithmic}
\usepackage{graphicx}
\usepackage{textcomp}
\usepackage{xcolor}
\usepackage{amsmath}
\def\BibTeX{{\rm B\kern-.05em{\sc i\kern-.025em b}\kern-.08em
    T\kern-.1667em\lower.7ex\hbox{E}\kern-.125emX}}
\begin{document}

\title{On the Transferability of Adversarial Examples between Encrypted Models}

\author{\IEEEauthorblockN{1\textsuperscript{st} Miki Tanaka}
\IEEEauthorblockA{\textit{Tokyo Metropolitan University} \\
Tokyo, Japan \\
tanaka-miki@ed.tmu.ac.jp}
\and
\IEEEauthorblockN{2\textsuperscript{nd} Isao Echizen}
\IEEEauthorblockA{\textit{National Institute of Informatics (NII)} \\
Tokyo, Japan \\
iechizen@nii.ac.jp}
\and
\IEEEauthorblockN{3\textsuperscript{rd} Hitoshi Kiya}
\IEEEauthorblockA{\textit{Tokyo Metropolitan University} \\
Tokyo, Japan \\
kiya@tmu.ac.jp}
}

\maketitle

\begin{abstract}
Deep neural networks (DNNs) are well known to be vulnerable to adversarial examples (AEs). In addition, AEs have adversarial transferability, namely, AEs generated for a source model fool other (target) models. In this paper, we investigate the transferability of models encrypted for adversarially robust defense for the first time. To objectively verify the property of transferability, the robustness of models is evaluated by using a benchmark attack method, called AutoAttack. In an image-classification experiment, the use of encrypted models is confirmed not only to be robust against AEs but to also reduce the influence of AEs in terms of the transferability of models. 
\end{abstract}


\section{Introduction}
Deep neural networks (DNNs) have been deployed in many applications including security-critical ones such as biometric authentication and automated driving, but DNNs are vulnerable to adversarial examples (AEs), which are perturbed by noise to mislead DNNs without affecting human perception. In addition, AEs generated for a source model fool other (target) models, and this property is called adversarial transferability. This transferability allows attackers to use a substitute model to generate AEs that may also fool other target models, so reducing the influence of the transferability has  become an urgent issue. Many studies have investigated both AEs and the transferability of AEs to build models robust against these attacks. In contrast, various methods for generating AEs have also been proposed to fool DNNs. 

One of the methods for constructing models robust against AEs is to train models by using encrypted images \cite{block_wise,Encryption_Inspired_Adversarial_Defense,ensemble}, which was inspired by learnable image encryption \cite{overview,Learnable_Image_Encryption,Block-wise_Scrambled_Image_Recognition,Encryption-Then-Compression_Systems,A_fast_image-scramble_method,Unitary_Transform-Based_Template_Protection,A_GAN-Based_Image_Transformation_Scheme}. The method was demonstrated to be robust against various AEs, but it has never been evaluated in terms of the transferability of AEs. Accordingly, in this paper, we aim to evaluate the method with encrypted images in terms of the transferability and encryption settings. In addition, the evaluation of encrypted models is carried out under the use of a benchmark attack method, referred to as AutoAttack, which was proposed to objectively evaluate the robustness of models against AEs. In an experiment, the use of encrypted models is verified not only to be robust against AEs but to also reduce the influence of the transferability between models.

\section{Related Work}
\subsection{Adversarial examples}
AEs are classified into three groups based on the knowledge of a particular model and training data available to the adversary: white-box, black-box, and gray-box. Under white-box settings \cite{FGSM,PGD,FAB}, the adversary has direct access to the model, its parameters, training data, and defense mechanism. However, the adversary does not have any knowledge on the model, except the output of the model in black-box attacks \cite{onepixelattack,square,nattack}.
Situated between white-box and black-box methods are gray-box attacks that imply that the adversary knows something about the system. With the development of AEs, numerous adversarial defenses have been proposed in the literature. Conventional defenses have been compared under a benchmark attack framework called AutoAttack\cite{autoattack}.
%入力信号に敵対的事例と呼ばれる人為的に作成された微小なノイズを，ニューラルネットワークモデルに加えることによって，意図的に予測結果を操作する敵対的事例攻撃がある．敵対的手法には，ホワイトボックス型，ブラックボックス型，グレーボックス型に分けられる．ホワイトボックス型は，攻撃対象となるモデルの構造やパラメータ，防御手法全てにアクセスできると仮定される．一方，ブラックボックス型とは，入出力のみにアクセスでき，攻撃対象のモデルにアクセスできない攻撃である．グレーボックス型とは，モデルに関する一部の情報のみにアクセス可能な攻撃である．さらに，敵対的事例はtarget型とnon-target型に分けられる．target型のノイズは特定のラベルに予測結果を誘導するように学習する．non-target型のノイズは正解のラベルの信頼度を下げるように学習される．

Many studies \cite{transferability_in_machine_learning,Intriguing_properties,Delving_into_Transferable_AE,On_the_robustness_of_ViT} have investigated adversarial transferability. The transferability in these studies is classified into two groups, i.e., non-targeted and targeted transferability, in accordance with the objective of the adversarial attack. However, no adversarial defense method with encrypted models has ever been evaluated in terms of robustness against adversarial transferability.
%あるモデルに対して作成された敵対的事例が，パラメータや構成が異なるモデルの予測結果も間違わせるという，Transfabilityと呼ばれる現象が報告されている[][]．ブラックボックス型の転送ベースの攻撃は，このTransfabilityを利用した攻撃である，
\subsection{AutoAttack}
Many defenses against AEs have been proposed, but it is very difficult to judge the value of defense methods without an independent test.
For this reason, AutoAttack\cite{autoattack}, which is an ensemble of adversarial attacks used to test adversarial robustness objectively, was proposed as a benchmark attack.
AutoAttack consists of four attack methods: Auto-PGD-cross entropy (APGD-ce)\cite{autoattack}, APGD-target (APGD-t), FAB-target (FAB-t)\cite{FAB}, and Square Attack\cite{square}, as summarized in Table \ref{tb:autoattack}.
In this paper, we use these four attack methods to objectively evaluate the transferability of AEs.
% 種々の敵対的攻撃と同様に，種々の防御法が提案されている．しかし，各手法の客観的比較は容易ではない．Autoattackは，防御手法の耐性を客観的に評価・比較するために提案された攻撃のアンサンブルである．Autoattackはホワイトボックス型とブラックボックス型を含む4つの攻撃手法で構成され，一つのモデルを異なる攻撃手法によって繰り返し攻撃する．本稿ではAutoattackを構成している攻撃手法の転移性を調査する．

\begin{table}[h]
    \caption{Attack methods used in AutoAttack}
    \label{tb:autoattack}
    \begin{center}
        \scalebox{1.0}{
    \begin{tabular}{c|c|c|}
        Attack & Target (T)/ Non-target (N) & White-box (W)/ Black-box (B)\\ \hline
        APGD-ce & N & W  \\ 
        APGD-t & T & W \\
        FAB-t & T & W \\
        Square & N & B
    \end{tabular}%
        }
    \end{center}
\end{table}

\section{Analysis of encrypted models}
In this paper, the robustness of encrypted models is evaluated under various settings. Targets for comparison are summarized here.

\subsection{Type of model}
Various models have been proposed for image classification tasks. The residual network (ResNet)\cite{ResNet} and very deep convolutional network (VGGNet)\cite{VGG} use a convolutional neural network(CNN).
In contrast, vision transformers (ViT) \cite{ViT} do not. In previous work, the transferability between CNN models and ViT was mentioned to be lower than the transferability between CNN models \cite{On_the_robustness_of_ViT}.
In this paper, we use three CNN models, ResNet18, ResNet50, and VGG16, and we also use ViT to investigate the transferability of AEs between models in addition to encrypted CNN models.

\subsection{Encrypted model}
\begin{figure}
    \centering
    \includegraphics[width=7cm]{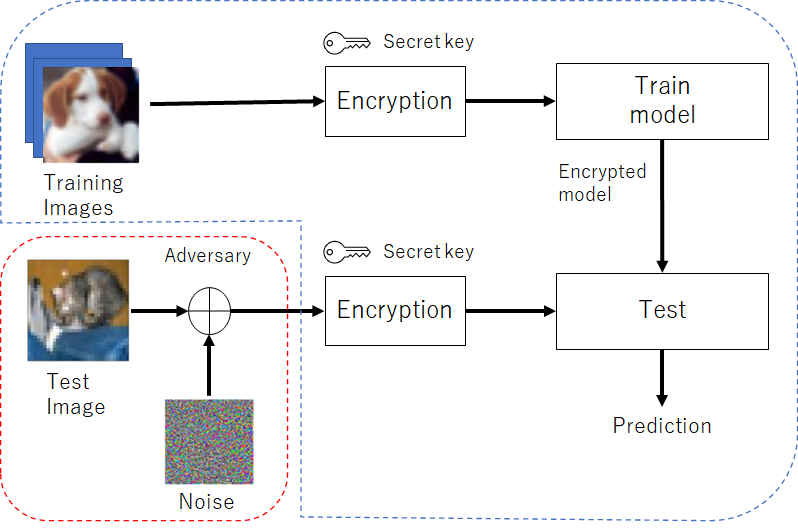}
    \caption{Adversarial defense with encrypted model}
    \label{fig:encrypted_model}
\end{figure}

\begin{figure}
    \centering
    \includegraphics[width=5cm]{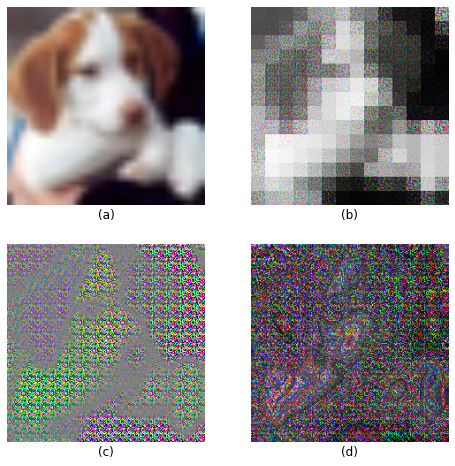}
    \caption{Example of transformed images (M=16) (a): plain image, (b): SHF, (c): NP, (d): FFX}
    \label{fig:transform_image}
\end{figure}

A block-wise transformation with secret keys was proposed for adversarial defense\cite{block_wise}, where a model is trained by using encrypted images as below (see Fig. \ref{fig:encrypted_model}).
\begin{enumerate}
    \item Each training image $x$ is divided into blocks with a size of $M \times M$.
    \item Every block in $x$ is encrypted by a transformation algorithm with secret keys to generate encrypted image.
    \item A model is trained by using the encrypted images to generate an encrypted model.
    \item A query image is encrypted with key $K$, and an encrypted image is then input to the encrypted model to get an estimation result.
\end{enumerate}
There are two parameters in steps 1) and 2), when encrypting a model: block size $M$ and transformation algorithm. In \cite{block_wise}, three transformation algorithms were proposed: pixel shuffling (SHF), bit flipping (NP), and format-preserving, Feistel-based encryption (FFX).
Figure \ref{fig:transform_image} shows an example of images encrypted from the original image with a size of 224 $\times$ 224 in Fig. \ref{fig:transform_image}(a) by using these three algorithms with $M=16$.

In this paper, we evaluate the transferability of AEs between models including encrypted ones.

\section{Experiment}
\subsection{Experiment setup}
In the experiment, we used four networks for image classification, ResNet18, ResNet50, VGG16, and ViT, to evaluate the transferability of AEs. In addition, ResNet18 was used for generating encrypted models where the above three transformation algorithms, SHF, NP, and FFX were applied to images in accordance with the steps in Sec I\hspace{-.1em}I\hspace{-.1em}I B.
The experiment was carried out on the CIFAR-10 dataset (with 10 classes), which consists of 60,000 color images (dimension of 3 $\times$ 32 $\times$ 32), where 50,000 of the images are for training and 10,000 for testing, and each class contains 6000 images. The images in the dataset were resized to 3 $\times$ 224 $\times$ 224 for fitting with pretrained ViT models.
AEs were generated by using four attack methods under the $l_\infty$ norm with $\epsilon = 8/255$ used in AutoAttack: APGD-ce, APGD-t, FAB-t, and Square. For white-box attacks, the adversary was assumed to be able to access the secret keys.
%敵対的事例はAutoAttackに含まれるAPGD-ce, APGD-t, FAB-t, Squareの4つの攻撃手法によって作成された．敵対者はResNet18, ViT, encrypted model (ResNet18, SHF, 4)を対象にこうげきした．但し，encryptedモデルを攻撃する時には敵対者は秘密鍵にアクセスできるものとする．

The transferability of AEs was evaluated by using two assessment criteria: accuracy (Acc) and attack success rate (ASR). Acc is given by
\begin{equation}
    \text{Acc} = \frac{100}{N} \sum_{k=1}^N \begin{cases} 
    1\  C_t(x_k^{adv})= y_k \\
    0\  \text{otherwise},
    \end{cases} 
\end{equation}
where $x^{adv}_k$ is an AE generated from an image $x_k$, $y_k$ is a label of $x_k$, $C_t$ is a target classifier, and $N$ is the number of input images. 
Acc is in the range of [0, 100] and a higher value indicates that images are classified more correctly. 

The ASR between a source classifier $C_s$ and $C_t$ is also given by

\begin{eqnarray}
    \text{ASR} = \frac{100}{N_c} \sum^{N_c}_{k=1} 
    \begin{cases} 
    1\  A_{C_t}(x_k,y_k) \land \{C_s(x_k)=y_k\} \\
    0\  \text{otherwise},
    \end{cases} \\
    A_C(x,y) = \{C(x)=y\} \land \{C(x^{adv})\neq y\},
\end{eqnarray}
where $N_c$ is the number of images correctly classified in both $C_t$ and $C_s$. ASR is in the range of [0,100], and a lower value indicates that the transferability is lower. 

\subsection{Results}
\subsubsection{Transferability between CNN and ViT}
%CNNとViTの転移性
As shown in Tables \ref{tb:Acc_CNN_ViT_ResNet} and \ref{tb:ASR_CNN_ViT_ResNet}, the transferability between plain models was evaluated in terms of ACC and ASR, respectively, where ResNet18 was chosen as the source model for which AEs were generated. From the tables, the AEs generated for the source model misled all CNN models including ResNet50 and VGG16. In contrast, they could not mislead the ViT models.

Tables \ref{tb:Acc_CNN_ViT_ViT} and \ref{tb:ASR_CNN_ViT_ViT} show the results obtained when ViT was chosen as the source model. The AEs generated with the ViT models could not fool the CNN models successfully. Thus, the transferability between CNN and ViT models was confirmed to be low as described in \cite{On_the_robustness_of_ViT}.

%%ResNet18
\begin{table}[h]
    \caption{Transferability between CNN and ViT models (Source: ResNet18, Acc)}
    \label{tb:Acc_CNN_ViT_ResNet}
    \begin{center}
        \scalebox{1.0}{
    \begin{tabular}{c||c|c|c|c|}
    Target & APGD-ce & APGD-t & FAB-t & Square \\ \hline
    ResNet18 & 0.00 & 0.00 & 0.36 & 0 \\
    ResNet50 & 4.84 & 30.64 & 92.69 & 74.20 \\
    VGG16 & 43.39 & 62.81 & 91.59 & 85.91 \\
    ViT & 68.92 & 93.95 & 98.99 & 95.26 \\
    \end{tabular}%
        }
    \end{center}
\end{table}

\begin{table}[h]
    \caption{Transferability between CNN and ViT models (Source: ResNet18, ASR)}
    \label{tb:ASR_CNN_ViT_ResNet}
    \begin{center}
        \scalebox{1.0}{
    \begin{tabular}{c||c|c|c|c|}
    Target & APGD-ce & APGD-t & FAB-t & Square \\ \hline
    ResNet18 & 100.00 & 100.00 & 99.62 & 100.00 \\
    ResNet50 & 97.08 & 68.97 & 1.35 & 21.74 \\
    VGG16 & 54.39 & 33.06 & 0.69 & 7.68 \\
    ViT & 32.16 & 5.49 & 0.04 & 4.09 \\
    \end{tabular}%
        }
    \end{center}
\end{table}

%%ViT
\begin{table}[h]
    \caption{Transferability between CNN and ViT models (Source: ViT, Acc)}
    \label{tb:Acc_CNN_ViT_ViT}
    \begin{center}
        \scalebox{1.0}{
    \begin{tabular}{c||c|c|c|c|}
    Target & APGD-ce & APGD-t & FAB-t & Square \\ \hline
    ResNet18 & 85.45 & 89.34 & 93.60 & 87.04 \\
    ResNet50 & 86.28 & 89.18 & 93.45 & 83.89 \\
    VGG16 & 84.22 & 85.90 & 91.93 & 87.85 \\
    ViT & 0.00 & 0.00 & 0.00 & 0.88 \\
    \end{tabular}%
        }
    \end{center}
\end{table}

\begin{table}[h]
    \caption{Ttransferability between CNN and ViT models (Source: ViT, ASR)}
    \label{tb:ASR_CNN_ViT_ViT}
    \begin{center}
        \scalebox{1.0}{
    \begin{tabular}{c||c|c|c|c|}
    Target & APGD-ce & APGD-t & FAB-t & Square \\ \hline
    ResNet18 & 9.60 & 5.56 & 0.55 & 8.16 \\
    ResNet50 & 8.60 & 5.79 & 0.57 & 11.49 \\
    VGG16 & 9.51 & 7.97 & 0.35 & 5.87 \\
    ViT & 100.00 & 100.00 & 100.00 & 99.11 \\
    \end{tabular}%
        }
    \end{center}
\end{table}

\subsubsection{Transferability between CNN and encrypted CNN models}
%秘密鍵を用いた分類器
Next, the transferability between CNN and encrypted CNN models was confirmed experimentally as shown in Tables \ref{tb:Acc_CNN_enc_ResNet} and \ref{tb:ASR_CNN_enc_ResNet}, where ResNet18 was chosen as the source model for generating AEs as well, and ResNe18 was also encrypted for model protection.
The use of encrypted models was verified to be effective for making the transferability of AEs weak. In particular, the use of FFX and a large block size enhanced the effect.

%%ブロック毎
\begin{table}[h]
    \caption{Transferability between CNN and encrypted models (Source: ResNet18, Target: Encrypted ResNet18, Acc)}
    \label{tb:Acc_CNN_enc_ResNet}
    \begin{center}
        \scalebox{1.0}{
    \begin{tabular}{c|c||c|c|c|c|}
    transform & block size & APGD-ce & APGD-t & FAB-t & Square \\ \hline
    SHF & 4 & 2.71 & 22.16 & 92.43 & 80.12 \\
    SHF & 8 & 6.25 & 37.42 & 93.00 & 76.61 \\
    SHF & 16 & 41.84 & 75.88 & 92.06 & 80.12 \\ \hline
    NP & 4 & 3.01 & 21.81 & 93.22 & 75.60 \\
    NP & 8 & 2.75 & 21.88 & 93.22 & 75.31 \\
    NP & 16 & 13.12 & 56.58 & 93.45 & 80.42 \\ \hline
    FFX & 4 & 18.93 & 68.58 & 91.91 & 84.40 \\
    FFX & 8 & 35.73 & 77.25 & 92.05 & 85.06 \\
    FFX & 16 & 71.22 & 86.02 & 92.29 & 85.69 \\
    \end{tabular}%
        }
    \end{center}
\end{table}

\begin{table}[h]
    \caption{Transferability between CNN and encrypted models (Source: ResNet18, Target: Encrypted ResNet18, ASR)}
    \label{tb:ASR_CNN_enc_ResNet}
    \begin{center}
        \scalebox{1.0}{
    \begin{tabular}{c|c||c|c|c|c|}
    transform & block size & APGD-ce & APGD-t & FAB-t & Square \\ \hline
    SHF & 4 & 99.29 & 78.16 & 1.41 & 15.17 \\
    SHF & 8 & 95.76 & 61.75 & 1.03 & 19.21 \\
    SHF & 16 & 56.04 & 18.57 & 0.62 & 14.16 \\ \hline
    NP & 4 & 99.22 & 78.77 & 1.11 & 20.42 \\
    NP & 8 & 99.38 & 78.64 & 1.13 & 20.87 \\
    NP & 16 & 88.08 & 40.74 & 0.58 & 14.95 \\ \hline
    FFX & 4 & 81.81 & 27.87 & 2.51 & 10.86 \\
    FFX & 8 & 63.34 & 18.11 & 1.98 & 9.83 \\
    FFX & 16 & 24.37 & 8.18 & 1.29 & 8.77 \\
    \end{tabular}%
        }
    \end{center}
\end{table}

\subsubsection{Transferability between encrypted models}
%%秘密鍵を用いた分類器を攻撃
The transferability between models encrypted under different conditions was evaluated as shown in Tables \ref{tb:Acc_CNN_enc_SHF} and \ref{tb:ASR_CNN_enc_SHF}, where the parameters used for encryption were block size, transformation algorithm, and secret key, and the source model was encrypted with a block size of 4 and SHF. As shown in the tables, when different parameters were used for each model encryption, the transferability between encrypted models was reduced. In particular, the use of FFX and a large block size can reduce the transferability more between models as well. 

\begin{table}[h]
    \caption{Transferability between encrypted models (Source: ResNet18 encrypted with SHF and M=4, Target: Encrypted ResNet18, Acc)}
    \label{tb:Acc_CNN_enc_SHF}
    \begin{center}
        \scalebox{0.85}{
    \begin{tabular}{c|c||c|c|c|c|}
    transform & block size & APGD-ce & APGD-t & FAB-t & Square \\ \hline
    no (plain) & - & 85.35 & 88.14 & 93.95 & 83.58 \\ \hline
    SHF (same key as target) & 4 & 0.00 & 0.00 & 0.18 & 0.00 \\
    SHF (different key) & 4 & 69.99 & 79.60 & 93.26 & 79.34 \\
    SHF & 8 & 81.7 & 85.26 & 93.83 & 78.89 \\
    SHF & 16 & 87.08 & 88.77 & 92.57 & 81.43 \\ \hline
    NP & 4 & 86.76 & 88.81 & 94.11 & 79.64 \\
    NP & 8 & 83.73 & 86.62 & 94.19 & 78.20 \\
    NP & 16 & 90.41 & 91.34 & 93.90 & 83.43 \\ \hline
    FFX & 4 & 90.31 & 92.03 & 93.21 & 86.40 \\
    FFX & 8 & 91.54 & 92.18 & 93.20 & 87.37 \\
    FFX & 16 & 90.60 & 91.15 & 92.74 & 87.02 \\
    \end{tabular}%
        }
    \end{center}
\end{table}

\begin{table}[h]
    \caption{Transferability between encrypted models (Source: ResNet18 encrypted with SHF and M=4, Target: Encrypted ResNet18, ASR)}
    \label{tb:ASR_CNN_enc_SHF}
    \begin{center}
        \scalebox{0.85}{
    \begin{tabular}{c|c||c|c|c|c|}
    transform & block size & APGD-ce & APGD-t & FAB-t & Square \\ \hline
    no (plain) & - & 9.70 & 6.73 & 0.16 & 11.66 \\ \hline
    SHF (same key as target) & 4 & 100.00 & 100.00 & 99.81 & 100 \\
    SHF (different key) & 4 & 25.84 & 15.28 & 0.27 & 15.79 \\
    SHF & 8 & 13.61 & 9.8 & 0.12 & 16.67 \\
    SHF & 16 & 6.58 & 4.7 & 0.06 & 12.92 \\ \hline
    NP & 4 & 8.37 & 6.14 & 0.14 & 16.02 \\
    NP & 8 & 11.88 & 8.66 & 0.19 & 17.89 \\
    NP & 16 & 4.17 & 3.16 & 0.09 & 11.88 \\ \hline
    FFX & 4 & 4.44 & 2.61 & 1.22 & 8.72 \\
    FFX & 8 & 2.99 & 2.23 & 1.04 & 7.53 \\
    FFX& 16 & 3.68 & 3.16 & 1.2 & 7.41 \\
    \end{tabular}%
        }
    \end{center}
\end{table}

\section{Conclusion}
In this paper, we investigated the transferability of models including encrypted ones. To objectively verify the transferability, four attack methods used in AutoAttack, were used to generate AEs from a source model.
In the experiment, the use of encrypted models was confirmed not only to be robust against AEs but to also reduce the influence of the transferability between models. In addition, the use of FFX and a large block for image encryption was effective for making the transferability of a model weak.

\section*{Acknowledgments}
The research was partially supported by JST CREST (Grant Number JPMJCR20D3) and ROIS NII Open Collaborative Research 2022-(22S1401).

%\bibliographystyle{IEEEtran}
%\bibliography{citation}

\end{document}